\pdfoutput=1

\documentclass[11pt]{article}

\usepackage[]{ACL2023}

\usepackage{times}
\usepackage{latexsym}

\usepackage[T1]{fontenc}

\usepackage[utf8]{inputenc}

\usepackage{microtype}

\usepackage{inconsolata}

\usepackage{times}
\usepackage{latexsym}
\usepackage{graphicx}
\usepackage{url}
\usepackage{CJK}
\usepackage{multirow}
\usepackage{multicol}
\usepackage{booktabs,colortbl}
\usepackage{pifont}
\usepackage{enumitem}
\usepackage{arydshln}
\usepackage{float}
\usepackage{caption}

\newcommand{\cmark}{\ding{51}}%
\newcommand{\xmark}{\ding{55}}%
\definecolor{forestgreen}{rgb}{0.13, 0.55, 0.13}

\title{A Survey on Zero Pronoun Translation}

\author{Longyue Wang\thanks{~~Longyue Wang and Siyou Liu contributed equally to this work.}~~, Siyou Liu$^{*}$, Mingzhou Xu, Linfeng Song, Shuming Shi, Zhaopeng Tu\\
  Tencent AI Lab \\
  \texttt{\{vinnylywang,lifengjin,shumingshi,zptu\}@tencent.com} \\
  \texttt{guofeng-ai@googlegroups.com} \\}

\begin{document}
\maketitle
\begin{abstract}
Zero pronouns (ZPs) are frequently omitted in pro-drop languages (e.g. Chinese, Hungarian, and Hindi), but should be recalled in non-pro-drop languages (e.g. English). This phenomenon has been studied extensively in machine translation (MT), as it poses a significant challenge for MT systems due to the difficulty in determining the correct antecedent for the pronoun. This survey paper highlights the major works that have been undertaken in zero pronoun translation (ZPT) after the neural revolution, so that researchers can recognise the current state and future directions of this field. We provide an organisation of the literature based on evolution, dataset, method and evaluation. In addition, we compare and analyze competing models and evaluation metrics on different benchmarks. We uncover a number of insightful findings such as: 1) ZPT is in line with the development trend of large language model; 2) data limitation cause learning bias in languages and domains; 3) performance improvements are often reported on single benchmarks, but advanced methods are still far from real-world use; 4) general-purpose metrics are not reliable on nuances and complexities of ZPT, emphasizing the necessity of targeted metrics; 5) apart from commonly-cited errors, ZPs will cause risks of gender bias.

\end{abstract}

\section{Introduction}
\label{sec:1}

Pronouns play an important role in natural language, as they enable speakers to refer to people, objects, or events without repeating the nouns that represent them. 
Zero pronoun (ZP)\footnote{ZP is also called dropped pronoun. The linguistic concept is detailed in Appendix \S\ref{app:a-3}.} is a complex phenomenon that appears frequently in pronoun-dropping (pro-drop) languages such as Chinese, Hungarian, and Hindi. Specifically, pronouns are often omitted when they can be pragmatically or grammatically inferable from intra- and inter-sentential contexts~\cite{li1979third}. 
Since recovery of such ZPs generally fails, this poses difficulties for several generation tasks, including dialogue modelling~\cite{su2019improving}, question answering~\cite{tan2021coupling}, and machine translation~\cite{wang2019discourse}.

When translating texts from pro-drop to non-pro-drop languages (e.g. Chinese$\Rightarrow$English), this phenomenon leads to serious problems for translation models in terms of: 1) \textit{completeness}, since translation of such invisible pronouns cannot be normally reproduced; 2) \textit{correctness}, because understanding the semantics of a source sentence needs to identifying and resolving the pronominal reference. 

\begin{CJK}{UTF8}{gbsn}
Figure~\ref{fig:zp_overview} shows ZP examples in three typological patterns determined by language family (detailed in Appendix \S\ref{app:a-1}). Taking a full-drop language for instance, the first-person subject and third-person object pronouns are omitted in Hindi input while these pronouns are all compulsory in English translation. This is not a problem for human beings since we can easily recall these missing pronoun from the context. However, even a real-life MT system still fails to accurately translate ZPs.
\end{CJK}

\begin{figure*}[t!]
    \centering
    \includegraphics[width=0.99\textwidth]{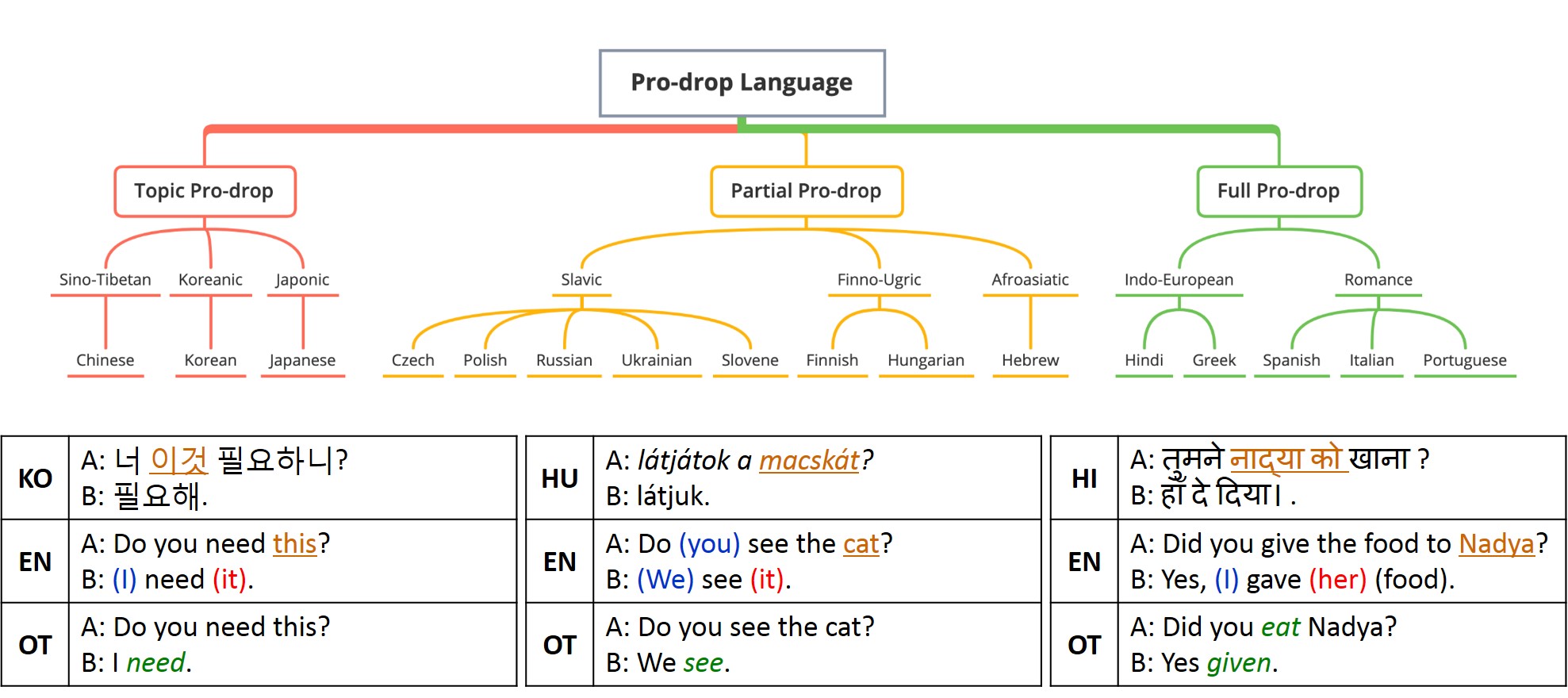}
    \caption{An overview of pro-drop languages by considering their typological patterns and language families. Example of ZP phenomenon in other languages (i.e. Korean, Hungarian  and Hindi). Words in brackets are pronouns that are invisible in source language ({\color{blue}implicit} and {\color{red}explicit}). The underlined words are corresponding antecedents. ``EN'' represents the human translation in English, which is a non-pro-drop language. ``OT'' is output translated by SOTA NMT systems with {\color{forestgreen} \it inappropriate translations}.} 
    \label{fig:zp_overview}
\end{figure*}

In response to this problem, zero pronoun translation (ZPT) has been studied extensively in the MT community on three significant challenges:
\begin{itemize} [leftmargin=*,topsep=0.1em,itemsep=0.1em,parsep=0.1em]

\item \textit{Dataset}: there is limited availability of ZP-annotated parallel data, making it difficult to develop systems that can handle ZP complexities.

\item \textit{Approach}: due to the ability to capture semantic information with distributed representations, ideally, the representations of NMT should embed ZP information by learning the alignments between bilingual pronouns from the training corpus. In practice, however, NMT models only manage to successfully translate some simple ZPs, but still fail when translating complex ones (e.g. subject vs. object ZPs).

\item \textit{Evaluation}: general evaluation metrics for MT are not sensitive enough to capture translation errors caused by ZPs.

\end{itemize}

We believe that it is the right time to take stock of what has been achieved in ZPT, so that researchers can get a bigger picture of where this line of research stands. In this paper, we present a survey of the major works on datasets, approaches and evaluation metrics that have been undertaken in ZPT.
We first introduce the background of linguistic phenomenon and literature selection in Section~\ref{sec:2}.
Section \ref{sec:3} discusses the evolution of ZP-related tasks.
Section~\ref{sec:4} summarizes the annotated datasets, which are significant to pushing the studies move forward. Furthermore, we investigated advanced approaches for improving ZPT models in Section~\ref{sec:5}. In addition to this, Section~\ref{sec:6} covers the evaluation methods that have been introduced to account for improvements in this field. We conclude by presenting avenues for future research in Section~\ref{sec:7}.

\section{Background}
\label{sec:2}

\subsection{Linguistic Phenomenon}
\label{sec:2.1}

\paragraph{Definition of Zero Pronoun}

Cohesion is a significant property of discourse, and it occurs whenever ``the interpretation of some element in the discourse is dependent on that of another'' \cite{halliday1976ruqaiya}. As one of cohesive devices, anaphora is the use of an expression whose interpretation depends specifically upon antecedent expression while zero anaphora is a more complex scenario in pro-drop languages. A ZP is a gap in a sentence, which refers to an entity that supplies the necessary information for interpreting the gap~\cite{zhao2007identification}. 
ZPs can be categorized into anaphoric and non-anaphoric ZP according to whether it refers to an antecedent or not.
In pro-drop languages such as Chinese and Japanese, ZPs occur much more frequently compared to non-pro-drop languages such as English. The ZP phenomenon can be considered one of the most difficult problems in natural language processing \cite{peral2003translation}.

\paragraph{Extent of Zero Pronoun}

\begin{CJK}{UTF8}{gbsn}
To investigate the extent of pronoun-dropping, we quantitatively analyzed ZPs in two corpora and details are shown in Appendix \S\ref{app:a-2}.
We found that the frequencies and types of ZPs vary in different genres: (1) 26\% of Chinese pronouns were dropped in the dialogue domain, while 7\% were dropped in the newswire domain; (2) the most frequent ZP in newswire text is the third person singular 它 (``it'') ~\cite{Baran2012AnnotatingDP}, while that in SMS dialogues is the first person 我 (``I'') and 我们 (``we'')~\cite{rao2015dialogue}. This may lead to differences in model behavior and quality across domains. 
This high proportion within informal genres such as dialogues and conversation shows the importance of addressing the challenge of translation of ZPs.
\end{CJK}

\subsection{Literature Selection}
\label{sec:2.2}

We used the following methodology to provide a comprehensive and unbiased overview of the current state of the art, while minimizing the risk of omitting key references:
\begin{itemize} [leftmargin=*,topsep=0.1em,itemsep=0.1em,parsep=0.1em]

\item \textit{Search Strategy}: We conducted a systematic search in major databases (e.g. Google Scholar) to identify the relevant articles and resources. Our search terms included combinations of keywords, such as "zero pronouns," "zero pronoun translation," and "coreference resolution." 

\item \textit{Selection Criteria}: To maintain the focus and quality of our review, we established the following criteria. (1) Inclusion, where articles are published in journals, conferences and workshop proceedings. (2) Exclusion, where articles that are not available in English or do not provide sufficient details to assess the validity of their results. 

\item \textit{Screening and Selection}: First, we screened the titles and abstracts based on our Selection Criteria. Then, we assessed the full texts of the remaining articles for eligibility. We also checked the reference lists of relevant articles to identify any additional sources that may have been missed during the initial search.

\item \textit{Data Extraction and Synthesis}: We extracted key information from the selected articles, such as dataset characteristics, and main findings. This data was synthesized and organized to provide a comprehensive analysis of the current state of the art in ZPT.

\end{itemize}

\section{Evolution of Zero Pronoun Modelling}
\label{sec:3}

\begin{figure*}[t]
    \centering
    \includegraphics[width=0.92\textwidth]{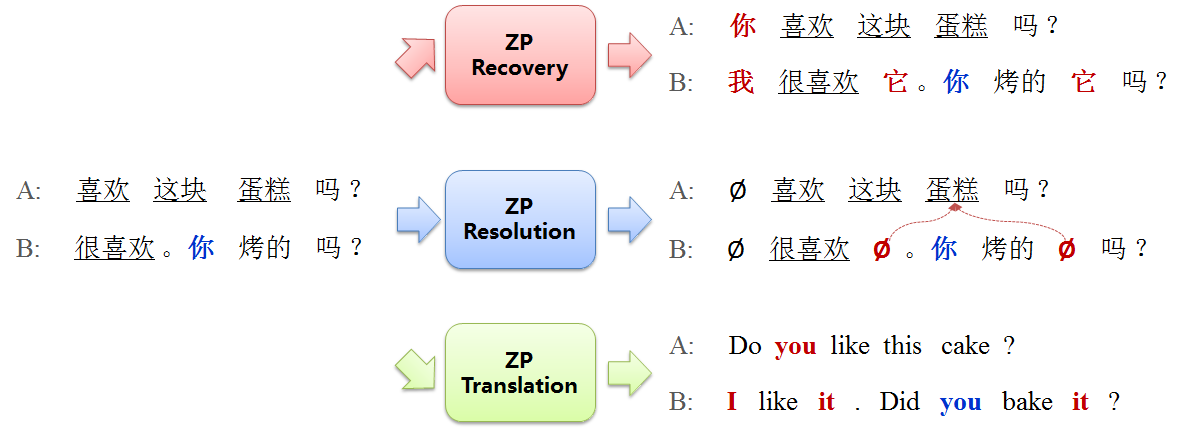}
    \caption{An overview of three ZP-aware tasks (taking Chinese-English for instance): ZP resolution, ZP recovery and ZP translation. As seen, the input is the same while the output varies according to different tasks.} 
    \label{fig:zp_tasks}
\end{figure*}

Considering the evolution of ZP modelling, we cannot avoid discussing other related tasks. Thus, we first review three typical ZP tasks and conclude their essential relations and future trends.

\subsection{Overview}
\label{sec:3-1}

ZP resolution is the earliest task to handle the understanding problem of ZP~\citep{zhao2007identification}. ZP recovery and translation aim to directly generate ZPs in monolingual and crosslingual scenarios, respectively~\citep{yang2010chasing,chung2010effects}.
This is illustrated in Figure~\ref{fig:zp_tasks}. 

\paragraph{Zero Pronoun Resolution}
The task contains three steps: {ZP} detection, anaphoricity determination and reference linking.
Earlier works investigated rich features using traditional ML models \cite{zhao2007identification,Kong:2010:EMNLP,chen2013chinese,chen2015chinese}. Recent studies exploited neural models to achieve the better performance~\cite{chen2016chinese,yin2018zero,song2020zpr2}. 
The CoNLL2011 and CoNLL2012\footnote{\url{https://cemantix.org}.} are commonly-used benchmarks on modeling unrestricted coreference. The corpus contains 144K coreference instances, but dropped subjects only occupy 15\%.

\paragraph{Zero Pronoun Recovery}

Given a source sentence, this aims to insert omitted pronouns in proper positions without changing the original meaning~\cite{yang2010chasing,yang2015recovering,yang2019recovering}. It is different from ZP resolution, which identifies the antecedent of a referential pronoun~\cite{mitkov2014anaphora}. Previous studies regarded ZP recovery as a classification or sequence labelling problem, which only achieve 40$\sim$60\% F1 scores on closed datasets~\cite{zhang2019neural,song2020zpr2}, indicating the difficulty of generating ZPs. It is worth noting that ZP recovery models can work for ZPT task in a {pipeline} manner: input sentences are labeled with ZPs using an external recovery system and then fed into a standard MT model~\cite{chung2010effects,wang2016naacl}.

\paragraph{Zero Pronoun Translation}

When pronouns are omitted in a source sentence, ZPT aims to generate ZPs in its target translation. Early studies have investigate a number of works for SMT models~\citep{chung2010effects,Nagard:2010:ACL,Taira:2012:SSSST,xiang2013enlisting, wang2016naacl}. Recent years have seen a surge of interest in NMT~\citep{yu2020better,wang2018translating}, since the problem still exists in advanced NMT systems.
ZPT is also related to pronoun translation, which aims to correctly translate explicit pronoun in terms of feminine and masculine. The DiscoMT\footnote{\url{https://aclanthology.org/W15-2500}.} is a commonly-cited benchmark on pronoun translation, however, there was no standard ZPT benchmarks up until now.

\subsection{Discussions and Findings}
\label{sec:3-2}

By comparing different ZP-aware tasks, we found three future trends:

\begin{itemize} [leftmargin=*,topsep=0.1em,itemsep=0.1em,parsep=0.1em]

\item[1.] {\bf From Intermediate to End}. In real-life systems, ZP resolution and recovery are intermediate tasks while ZPT can be directly reflected in system output. ZP resolution and recovery will be replaced by ZPT although they currently work with some MT systems in a pipeline way.

\item[2.] {\bf From Separate To Unified}. With the development of large language models (LLMs), it is unnecessary to keep a specific model for each task. 
For example, \citet{song2020zpr2} leveraged a unified BERT-based architecture to model ZP resolution and recovery.
Furthermore, we observed that ChatGPT\footnote{\url{https://openai.com/blog/chatgpt}.} already possesses the capability for ZP resolution and recovery.
\end{itemize}

\begin{table*}[t]
    \centering
    \scalebox{0.93}{
    \begin{tabular}{l c c lr c c c}
    \toprule
    \multirow{2}{*}{\bf Dataset} & \multirow{2}{*}{\bf Lang.} & \multirow{2}{*}{\bf Anno.} & \multirow{2}{*}{\bf Domain} & \multirow{2}{*}{\bf Size} & \multicolumn{3}{c}{\bf Task} \\
    \cmidrule(lr){6-8}
    &&&&&  {Reso.} & {Reco.} & {Trans.} \\
    \midrule
    OntoNotes~{\small\cite{pradhan2012conll}} & ZH & Human & Mixed Sources & 42.6K & {\color{red}\cmark} & {\color{blue}\xmark} & {\color{blue}\xmark} \\
    OntoNotes~{\small\cite{aloraini2020cross}} & AR & Human & News & 9.4K & {\color{red}\cmark} & {\color{blue}\xmark} & {\color{blue}\xmark} \\
    CTB~{\small\cite{yang2010chasing}} & ZH & Human & News & 10.6K &   {\color{blue}\xmark} &   {\color{red}\cmark}& {\color{blue}\xmark}\\
    KTB~{\small\cite{chung2010effects}} & KO & Human & News & 5.0K &   {\color{blue}\xmark} &   {\color{red}\cmark}& {\color{blue}\xmark}\\
    BaiduKnows~{\small\cite{zhang2019neural}} & ZH & Human & Baidu Knows & 5.0K& {\color{blue}\xmark} & {\color{red}\cmark} & {\color{blue}\xmark} \\
    TVsub~{\small\cite{wang2018translating}} & ZH, EN & Auto & Movie Subtitles & 2.2M & {\color{blue}\xmark}  & {\color{blue}\xmark} & {\color{red}\cmark} \\
    ZAC~{\small\cite{pereira2009zac}} & PT & Human & Mixed Sources & 0.6K & {\color{red}\cmark} & {\color{blue}\xmark} & {\color{blue}\xmark} \\
    Nagoya~{\small\cite{zhan2015automatic}} & JA & Auto & Scientific Paper & 1.2K & {\color{red}\cmark} & {\color{blue}\xmark} & {\color{blue}\xmark}\\
    SKKU~{\small\cite{park2015zero}} & KO & Human & Dialogue & 1.1K & {\color{red}\cmark} & {\color{blue}\xmark} & {\color{blue}\xmark}\\
    UPENN~{\small\cite{prasad2000corpus}} & HI &  Human & News & 2.2K & {\color{red}\cmark} & {\color{blue}\xmark} & {\color{blue}\xmark} \\
    LATL~{\small\cite{russo2012italian}} & IT, ES & Human & Europarl & 2.0K & {\color{red}\cmark} & {\color{blue}\xmark} & {\color{red}\cmark} \\
    UCFV~{\small\cite{bacolini2017exploring}} & HE & Human & Dialogue & 0.1K & {\color{red}\cmark} & {\color{blue}\xmark} & {\color{blue}\xmark}\\
    \bottomrule  
    \end{tabular}}
    \caption{A summary of existing datasets regarding ZP. We classify them according to language (Lang.), annotation type (Anno.) and text domain. We also report the number of sentences (Size). ``Reso.'', ``Reco.'' and ``Trans.'' indicate whether a dataset can be used for specific ZP tasks. The symbol {\color{red}\cmark} or {\color{blue}\xmark} means ``Yes'' or ``No''.}
    \label{tab:copora}
\end{table*}

\section{Datasets}
\label{sec:4}

\subsection{Overview}
\label{sec:4-1}

Modeling ZPs has so far not been extensively explored in prior research, largely due to the lack of publicly available data sets. Existing works mostly focused on human-annotated, small-scale and single-domain corpora such as OntoNotes~\cite{pradhan2012conll,aloraini2020cross} and Treebanks~\cite{yang2010chasing,chung2010effects}. 
We summarize representative corpora as:
\begin{itemize} [leftmargin=*,topsep=0.1em,itemsep=0.1em,parsep=0.1em]

\item \textit{OntoNotes.}\footnote{\url{https://catalog.ldc.upenn.edu/LDC2013T19}.} This is annotated with structural information (e.g. syntax and predicate argument structure) and shallow semantics (e.g. word sense linked to an ontology and coreference). It comprises various genres of text (news, conversational telephone speech, weblogs, usenet newsgroups, broadcast, talk shows) in English, Chinese, and Arabic languages. ZP sentences are extracted for ZP resolution task~\cite{chen2013chinese,chen2016chinese}.

\item \textit{TVSub.}\footnote{\url{https://github.com/longyuewangdcu/tvsub}.} This extracts Chinese--English subtitles from television episodes. Its source-side sentences are automatically annotated with ZPs by a heuristic algorithm~\cite{wang2016naacl}, which was generally used to study dialogue translation and zero anaphora phenomenon~\cite{wang2018translating,tan2021coupling}.

\item \textit{CTB.}\footnote{\url{https://catalog.ldc.upenn.edu/LDC2013T21}.} This is a part-of-speech tagged and fully bracketed Chinese language corpus. The text are extracted from various domains including newswire, government documents, magazine articles, various broadcast news and broadcast conversation programs, web newsgroups and weblogs. Instances with empty category are extracted for ZP recovery task~\cite{yang2010chasing,chung2010effects}.

\item \textit{BaiduKnows.} The source-side sentences are collected from the Baidu Knows website,\footnote{\url{https://zhidao.baidu.com}.} which were annotated with ZP labels with boundary tags. It is widely-used the task of ZP recovery~\cite{zhang2019neural,song2020zpr2}.

\end{itemize}

\subsection{Discussions and Findings}
\label{sec:4-2}

\begin{CJK}{UTF8}{gbsn}
Table~\ref{tab:copora} lists statistics of existing ZP datasets and we found the limitations and trends:
\begin{itemize} [leftmargin=*,topsep=0.1em,itemsep=0.1em,parsep=0.1em]

\item[1.] {\bf Language Bias}. Most works used Chinese and Japanese datasets as testbed for training ZP models~\cite{song2020zpr2,ri2021zero}. However, there were limited data available for other pro-drop languages (e.g. Portuguese and Spanish), resulting that linguists mainly used them for corpus analysis~\cite{pereira2009zac,russo2012italian}. However, ZP phenomenon may vary across languages in terms of word form, occurrence frequency and category distribution, leading to learning bias on linguistic knowledge. Thus, it is necessary to establish ZP datasets for various languages~\cite{prasad2000corpus,bacolini2017exploring}.


\item[2.] {\bf Domain Bias}. Most corpora were established in one single domain (e.g. news), which may not contain rich ZP phenomena. Because the frequencies and types of ZPs vary in different genres~\cite{yang2015recovering}. Future works need more multi-domain datasets to better model behavior and quality for real-life use.

\item[3.] {\bf Become An Independent Research Problem}. Early works extracted ZP information from closed annotations (e.g. OntoNotes and Treebanks)~\cite{yang2010chasing,chung2010effects}, which were considered as a sub-problem of coreference or syntactic parsing. With further investigation on the problem, MT community payed more attention to it by manually or automatically constructing ZP recovery and translation datasets (e.g. BaiduKnows and TVsub)~\cite{wang2018translating,zhang2019neural}. 

\item[4.] {\bf Coping with Data Scarcity}. The scarcity of ZPT data remains a core issue (currently only 2.2M $\sim$ 0.1K sentences) due to two challenges: (1) it requires experts for both source ZP annotation and target translation~\cite{wang2016automatic,wang2018translating}; (2) annotating the training data manually spends much time and money. Nonetheless, it is still necessary to establish testing datasets for validating/analyzing the model performance. Besides, pre-trained modes are already equipped with some capabilities on discourse~\cite{chen2019evaluation,koto2021discourse}. This highlights the importance of formulating the downstream task in a manner that can effectively leverage the capabilities of the pre-trained models. 
\end{itemize}

\end{CJK}

\section{Approaches}
\label{sec:5}

\subsection{Overview}
\label{sec:5-1}

Early researchers have investigated several approaches for conventional statistical machine translation (SMT)~\cite{Nagard:2010:ACL,xiang2013enlisting,wang2016naacl}.
Modeling ZPs for advanced NMT models, however, has received more attention, resulting in better performance in this field~\cite{wang2018translating,tan2021coupling,hwang2021contrastive}.
Generally prior works fall into three categories: (1) {\bf Pipeline}, where input sentences are labeled with ZPs using an external ZP recovery system and then fed into a standard MT model~\cite{chung2010effects,wang2016naacl}; (2) {\bf Implicit}, where ZP phenomenon is implicitly resolved by modelling document-level contexts~\cite{yu2020better,ri2021zero}; (3) {\bf End-to-End}, where ZP prediction and translation are jointly learned in an end-to-end manner~\cite{wang2019one,tan2021coupling}.

\paragraph{Pipeline} 
\label{pip}

The pipeline method of ZPT borrows from that in pronoun translation~\cite{Nagard:2010:ACL,pradhan2012conll} due to the strong relevance between the two tasks.
\citet{chung2010effects} systematically examine the effects of empty category (EC)\footnote{In linguistics, it is an element in syntax that does not have any phonological content and is therefore unpronounced.} on SMT with pattern-, {CRF}- and parsing-based methods. The results show that this can really improve the translation quality, even though the automatic prediction of {EC} is not highly accurate. Besides, \citet{wang2016naacl,wang2016dropped,wang2017novel} proposed to integrate neural-based ZP recovery with SMT systems, showing better performance on both ZP recovery and overall translation.
When entering the era of NMT, ZP recovery is also employed as an external system.
Assuming that no-pro-drop languages can benefit pro-drop ones, \citet{ohtani2019context} tagged the coreference information in the source language, and then encoded it using a graph-based encoder integrated with NMT model. 
\citet{tan2019detecting} recovered ZP in the source sentence via a BiLSTM--CRF model~\cite{lample2016neural}. Different from the conventional ZP recovery methods, the label is the corresponding translation of ZP around with special tokens. They then trained a NMT model on this modified data, letting the model learn the copy behaviors.
\citet{tan2021coupling} used ZP detector to predict the ZP position and inserted a special token. Second, they used a attention-based ZP recovery model to recover the ZP word on the corresponding ZP position.

\begin{table*}[t]
    \centering
    \scalebox{0.95}{
    \begin{tabular}{l rr rr rr}
    \toprule
    \multirow{2}{*}{\bf Model} & \multicolumn{2}{c}{\bf TVsub} & \multicolumn{2}{c}{\bf BaiduKnows} & \multicolumn{2}{c}{\bf Webnovel}\\
    \cmidrule(lr){2-3} \cmidrule(lr){4-5} \cmidrule(lr){6-7}
    & BLEU & APT & BLEU & APT & BLEU & APT\\
    \midrule
    Baseline \cite{Vaswani:2017:NIPS} & 29.4 & 47.4 & 12.7 & 25.4 & 11.7 & 30.9\\
    \midrule
    Pipeline \cite{song2020zpr2} & 29.8 & 49.5 & 13.2 & 56.4 & 11.6 & 32.0\\
    Implicit \cite{ma2020simple} & 29.8 & 53.5 & 13.9 & 26.3 & 12.2 & 35.3\\
    End-to-End \cite{wang2018translating} & 30.0 & 52.3 & 12.3 & 30.4 & 12.0 & 33.4\\
    \midrule
    ORACLE & 32.8 & 86.9 & 14.7 & 88.8 & 12.8 & 85.1\\
    \bottomrule
\end{tabular}}
    \caption{A comparison of representative ZPT methods with different benchmarks. The ZPT methods are detailed in Section~\ref{sec:5-1}. The Baseline is a standard Transformer-big model while ORACLE is manually recovering ZPs in input sentences and then feeding them into the Baseline~\cite{wu2020tencent}. As detailed in Section~\ref{sec:4-1}, TVSub (both translation and ZP training data) and BaiduKnows (ZP training data) are widely-used benchmarks in movie subtitle and Q\&A forum domains, respectively. The Webnovel is our in-house testing data (no training data) in web fiction domain. As detailed in Section~\ref{sec:6-1}, BLEU is a general-purpose evaluation metric while APT is a ZP-targeted one.}
    \label{tab:ZPT_method}
\end{table*}

\paragraph{End-to-End}
Due the lack of training data on ZPT, a couple of studies pay attention to data augmentation. \citet{sugiyama2019data} employed the back-translation on a context-aware NMT model to augment the training data. With the help of context, the pronoun in no-pronoun-drop language can be translated correctly into pronoun-drop language. They also build a contrastive dataset to filter the pseudo data. Besides, \citet{kimura2019selecting} investigated the selective standards in detail to filter the pseudo data. \citet{ri2021zero} deleted the personal pronoun in the sentence to augment the training data. And they trained a classifier to keep the sentences that pronouns can be recovered without any context.

About model architecture, \citet{wang2018translating} first proposed a reconstruction-based approach to reconstruct the ZP-annotated source sentence from the hidden states of either encoder or decoder, or both. The central idea behind is to guide the corresponding hidden states to embed the recalled source-side ZP information and subsequently to help the NMT model generate the missing pronouns with these enhanced hidden representations.
Although this model achieved significant improvements, there nonetheless exist two drawbacks: 1) there is no interaction between the two separate reconstructors, which misses the opportunity to exploit useful relations between encoder and decoder representations; and 2) testing phase needs an external ZP prediction model and it only has an accuracy of 66\% in F1-score, which propagates numerous errors to the translation model. Thus, \citet{wang2018learning} further proposed to improve the reconstruction-based model by using {\em shared} reconstructor and joint learning. Furthermore, relying on external ZP models in decoding makes these approaches unwieldy in practice, due to introducing more computation cost and complexity. 

About learning objective, contrastive learning is often used to let the output more close to golden data while far away from negative samples. 
\citet{yang2019reducing} proposed a contrastive learning to reduce the word omitted error. To construct the negative samples, they randomly dropped the word by considering its frequency or part-of-speech tag. \citet{hwang2021contrastive} further considered the coreference information to construct the negative sample. According to the coreference information, they took place the antecedent in context with empty, mask or random token to get the negative samples. Besides, \citet{jwalapuram2020pronoun} served the pronoun mistranslated output as the negative samples while golden sentences as positive sample. To get the negative samples, they aligned the word between model outputs and golden references to get the sentences with mistranslated pronoun.

\paragraph{Implicit} 
\label{round} 
Some works consider not just the ZPT issue but rather focus on the overall discourse problem. The document-level NMT models~\cite{wang2017exploiting,werlen2018document,ma2020simple,lopes2020document} are expected to have strong capabilities in discourse modelling such as translation consistency and ZPT.
Another method is the round-trip translation, which is commonly-used in automatic post-editing (APE)~\cite{freitag2019ape}, quality estimation (QE)~\cite{moon2020revisiting} to correct of detect the translation errors. \citet{voita2019context} served this idea on context-aware NMT to correct the discourse error in the output. They employed the round-trip translation on monolingual data to get the parallel corpus in the target language. They then used the corpus to train a model to repair discourse phenomenon in MT output. \citet{wang2019one} proposed a fully unified ZPT model, which absolutely released the reliance on external ZP models at decoding time. Besides, they exploited to jointly learn inter-sentential context~\cite{Sordoni2015A} to further improve ZP prediction and translation.

\subsection{Discussions and Findings}
\label{sec:5-2}

Table~\ref{tab:copora} shows that only the TVsub is suitable for both training and testing in ZPT task, while others like LATL is too small and only suitable for testing. To facilitate fair and comprehensive comparisons of different models across different benchmarkss, we expanded the BaiduKnows by adding human translations and included in-house dataset\footnote{The Webnovel testing dataset contains 1,658 Chinese-English sentence pairs in 24 documents, with the target side translated by professional human translators.}. 
As shown in Table~\ref{tab:ZPT_method}, we re-implemented three representative ZPT methods and conducted experiments on three benchmarks, which are diverse in terms of domain, size, annotation type, and task. As the training data in three benchmarks decrease, the difficulty of modelling ZPT gradually increases.


\begin{CJK}{UTF8}{gbsn}
\begin{itemize} [leftmargin=*,topsep=0.1em,itemsep=0.1em,parsep=0.1em]

\item[1.] {\bf Existing Methods Can Help ZPT But Not Enough}. Three ZPT models can improve ZP translation in most cases, although there are still considerable differences among different domain of benchmarks (BLEU and {APT} $\uparrow$). Introducing ZPT methods has little impact on BLEU score (-0.4$\sim$+0.6 point on average), however, they can improve {APT} over baseline by +1.1$\sim$+30.1. When integrating golden ZP labels into baseline models (ORACLE), their BLEU and {APT} scores largely increased by +3.4 and +63.4 points, respectively. The performance gap between Oracle and others shows that there is still a large space for further improvement for ZPT. 

\item[2.] {\bf Pipeline Methods Are Easier to Integrate with NMT}. This is currently a simple way to enhance ZPT ability in real-life systems. As shown in Table~\ref{tab:zpr_case}, we analyzed the outputs of pipeline method and identify challenges from three perspectives: (1) {\em out-of-domain}, where it lacks in-domain data for training robust ZP recovery models. The distribution of ZP types is quite different between ZP recovery training data (out-of-domain) and ZPT testset (in-domain). This leads to that the ZP recovery model often predicts wrong ZP forms (possessive adjective vs. subject). (2) {\em error propagation}, where the external ZP recovery model may provide incorrect ZP words to the followed NMT model. As seen, \textsc{Zpr+} performs worse than a plain NMT model \textsc{Nmt} due to wrong pronouns predicted by the \textsc{Zpr} model (你们 vs. 我). (3) {\em multiple ZPs}, where there is a 10\% percentage of sentences that contain more than two ZPs, resulting in more challenges to accurately and simultaneously predict them. As seen, two ZPs are incorrectly predicted into ``我'' instead of ``他''.

\item[3.] {\bf Data-Level Methods Do Not Change Model Architecture}. This is more friendly to NMT. Some researchers targeted making better usage of the limited training data~\cite{tan2019detecting,ohtani2019context,tan2021coupling}. They trained an external model on the ZP data to recover the ZP information in the input sequence of the MT model~\cite{tan2019detecting,ohtani2019context,tan2021coupling} or correct the errors in the translation outputs~\cite{voita2019context}. Others aimed to up-sample the training data for the ZPT task~\cite{sugiyama2019data,kimura2019selecting,ri2021zero}. They preferred to improve the ZPT performance via a data augmentation without modifying the MT architecture~\cite{wang2016naacl,sugiyama2019data}. \citet{kimura2019selecting,ri2021zero} verified that the performance can be further improved by denoising the pseudo data. 

\item[4.] {\bf Multitask and Multi-Lingual Learning}. ZPT is a hard task to be done alone, researchers are investigating how to leverage other related NLP tasks to improve ZPT by training models to perform multiple tasks simultaneously~\cite{wang2018translating}. Since ZPT is a cross-lingual problem, researchers are exploring techniques for training models that can work across multiple languages, rather than being limited to a single language~\cite{aloraini2020cross}.

\end{itemize}
\end{CJK}

\begin{CJK}{UTF8}{gbsn}
\begin{table}[t]
    \centering
    \scalebox{0.92}{
    \begin{tabular}{c l l}
    \toprule
    \multirow{4}{*}{\rotatebox[origin=c]{90}{\tiny\bf 1. Out-of-Domain}} & \textsc{Inp.} & \textcolor{red}{[他的]$_{p}$} 主要 研究 领域 为 ...\\
    & \textsc{Nmt} & The main research areas are ...\\
    \cdashline{2-3}\noalign{\vskip 0.5ex}
    & \textsc{Zpr} & \textcolor{forestgreen}{我} 主要 研究 领域 为 ...\\
    & \textsc{Zpr+} & \textcolor{blue}{My} main research areas are ...\\
    \midrule
    \multirow{4}{*}{\rotatebox[origin=c]{90}{\tiny\bf 2. Error Propagation}} & \textsc{Inp.} & 如果 \textcolor{red}{[你们]$_{s}$} 见到 她 ...  \\[0.5ex]
    & \textsc{Nmt} & If \textcolor{blue}{you} see her ...\\
    \cdashline{2-3}\noalign{\vskip 0.5ex}
    & \textsc{Zpr} & 如果 \textcolor{forestgreen}{我} 见到 她 ...\\
    & \textsc{Zpr+} & If \textcolor{blue}{I} see her ...\\
    \midrule
    \multirow{4}{*}{\rotatebox[origin=c]{90}{\tiny\bf 3. Multiple ZPs}} & \textsc{Inp.} & \textcolor{red}{[他]$_{s}$} 好久没 ... \textcolor{red}{[他]$_{s}$} 怪 想念 的。\\[0.5ex]
    & \textsc{Nmt} & for a long time did not ... strange miss.\\
    \cdashline{2-3}\noalign{\vskip 0.5ex}
    & \textsc{Zpr} & \textcolor{forestgreen}{我} 好久没 ... \textcolor{forestgreen}{我} 怪 想念 的。\\
    & \textsc{Zpr+} &  \textcolor{blue}{I} haven't ... for a long time, \textcolor{blue}{I} miss.\\
    \bottomrule
    \end{tabular}}
    \caption{Errors in a pipeline-based ZPT and NMT models. \textsc{Inp.} represents Chinese input and \textsc{Nmt} indicates a sentence-level NMT models. \textsc{Zpr} denotes ZP-annotated output predicted by ZP recovery models. Red words are ZPs that are invisible in decoding.}
    \label{tab:zpr_case}
\end{table}
\end{CJK}

\section{Evaluation Methods}
\label{sec:6}
\subsection{Overview}
\label{sec:6-1}

There are three kinds of automatic metrics to evaluate performances of related models: 
\begin{itemize} [leftmargin=*,topsep=0.1em,itemsep=0.1em,parsep=0.1em]

\item {\em Accuracy of ZP Recovery}: this aims to measure model performance on detecting and predicting ZPs of sentences in one pro-drop language. For instance, the micro F1-score is used to evaluating Chinese ZPR systems~\citet{song2020zpr2}.\footnote{\url{https://github.com/freesunshine0316/lab-zp-joint}.}

\item {\em General Translation Quality}: there are a number of automatic evaluation metrics for measuring general performance of MT systems~\cite{snover2006study}. BLEU~\cite{papineni2002bleu} is the most widely-used one, which measures the precision of n-grams of the MT output compared to the reference, weighted by a brevity penalty to punish overly short translations. METEOR~\cite{meteor} incorporates semantic information by calculating either exact match, stem match, or synonymy match. Furthermore, COMET~\cite{rei2020comet} is a neural framework for training multilingual MT evaluation models which obtains new SOTA levels of correlation with human judgements.

\item {\em Pronoun-Aware Translation Quality}: Previous works usually evaluate ZPT using the BLEU metric~\cite{wang2016naacl,wang2018translating,yu2020better,ri2021zero}, however, general-purpose metrics cannot characterize the performance of ZP translation. As shown in Table~\ref{tab:zpr_case}, the missed or incorrect pronouns may not affect BLEU scores but severely harm true performances. To fix this gap, some works proposed pronoun-targeted evaluation metrics~\cite{werlen2017validation,laubli2018has}. 

\end{itemize}

\begin{table}[t]
    \centering
    \scalebox{0.95}{
    \begin{tabular}{l r r r r}
    \toprule
    \bf Metric & \bf T.S. & \bf B.K. & \bf I.H. & \bf Ave.\\
    \midrule
    BLEU & 0.09 & 0.76 & 0.57 & 0.47\\
    TER & 0.41 & 0.01 & 0.26 & 0.23\\
    METEOR & 0.23 & 0.74 & 0.28 & 0.42\\
    COMET & 0.59 & 0.15 & 0.37 & 0.37\\
    \midrule
    APT & \bf 0.68 & \bf 0.76 & \bf 0.58 & \bf 0.67\\
    \bottomrule
    \end{tabular}}
    \caption{Correlation between the manual evaluation and other automatic metrics, which are applied on different ZPT benchmarks, which are same as in Table~\ref{tab:ZPT_method}.}
    \label{tab:correlation}
\end{table}

\subsection{Discussions and Findings}
\label{sec:6-2}

As shown in Table~\ref{tab:correlation}, we compare different evaluation metrics on ZPT systems. About general-purpose metrics, we employed BLEU, TER, METEOR and COMET. About ZP-targeted metrics, we implemented and adapted {APT}~\cite{werlen2017validation} to evaluate ZPs, and experimented on three Chinese-English benchmarks (same as Section~\ref{sec:5-2}). For human evaluation, we randomly select a hundred groups of samples from each dataset, each group contains an oracle source sentence and the hypotheses from six examined MT systems. We asked expert raters to score all of these samples in 1 to 5 scores to reflect the cohesion quality of translations (detailed in Appendix \S\ref{app:a-4}). {The professional annotators are bilingual professionals with expertise in both Chinese and English. They have a deep understanding of the ZP problem and have been specifically trained to identify and annotate ZPs accurately.}
Our main findings are:

\begin{itemize} [leftmargin=*,topsep=0.1em,itemsep=0.1em,parsep=0.1em]

\item[1.] {\bf General-Purpose Evaluation Are Not Applicable to ZPT}. As seen, APT reaches around 0.67 Pearson scores with human judges, while general-purpose metrics reach 0.47$\sim$23. The APT shows a high correlation with human judges on three benchmarks, indicating that (1) general-purpose metrics are not specifically designed to measure performance on ZPT; (2)  researchers need to develop more targeted evaluation metrics that are better suited to this task.

\item[2.] {\bf Human Evaluations Are Required as A Complement}. Even we use targeted evaluation, some nuances and complexities remain unrecognized by automatic methods. Thus, we call upon the research community to employ human evaluation according to WMT~\cite{kocmi2022findings} especially in chat and literary shared tasks~\cite{farinha2022findings,wang2023literary}.

\item[3.] {\bf The Risk of Gender Bias}. The gender bias refers to the tendency of MT systems to produce output that reflects societal stereotypes or biases related to gender~\cite{vanmassenhove2019lost}. We found gender errors in ZPT outputs, when models make errors in identifying the antecedent of a ZP. This can be caused by the biases present in the training data, as well as the limitations in the models and the evaluation metrics. Therefore, researchers need to pay more attention to mitigate these biases, such as using diverse data sets and debiasing techniques, to improve the accuracy and fairness of ZPT methods.

\end{itemize}

\begin{figure}[t] 
    \centering
    \includegraphics[width=0.46\textwidth]{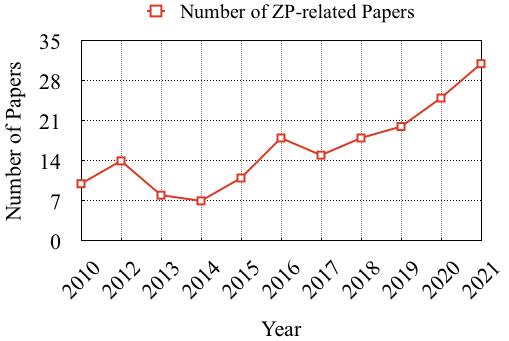}
    \caption{Number of papers mentioning ``zero pronoun'' per year according Google Scholar.} 
    \label{fig:paper}
\end{figure}

\section{Conclusion and Future Work}
\label{sec:7}

ZPT is a challenging and interesting task, which needs abilities of models on discourse-aware understanding and generation. 
Figure~\ref{fig:paper} best illustrates the increase in scientific publications related to ZP over the past few years.
This paper is a literature review of existing research on zero pronoun translation, providing insights into the challenges and opportunities of this area and proposing potential directions for future research.

As we look to the future, we intend to delve deeper into the challenges of ZPT. Our plan is to leverage large language models, which have shown great potential in dealing with complex tasks, to tackle this particular challenge \cite{lu2023error,wang2023document,lyu2023new}. Moreover, we plan to evaluate our approach on more discourse-aware tasks. Specifically, we aim to utilize the GuoFeng Benchmark~\cite{xu2022guofeng,wangguofeng}, which presents a comprehensive testing ground for evaluating the performance of models on a variety of discourse-level translation tasks. By doing so, we hope to gain more insights into the strengths and weaknesses of our approach, and continually refine it to achieve better performance.

\section*{Acknowledgement}
\label{sec:8}

The authors express their sincere gratitude to all reviewers whose keen interest and insightful feedback have significantly improved the quality of this paper. Their affirmation and encouragement have further solidified our commitment to the path of computational linguistics. This work is part of the {\em GuoFeng AI} ({guofeng-ai@googlegroups.com}) and {\em TranSmart} \cite{huang2021transmart} projects.

\section*{Limitations}

We list the main limitations of this work as follows:
\begin{itemize}[leftmargin=*,topsep=0.1em,itemsep=0.1em,parsep=0.1em]
    \item[1.] {\em Zero Pronoun in Different Languages}: The zero pronoun phenomenon may vary across languages in terms of word form, occurrence frequency and category distribution etc. Due to page limitation, some examples are mainly discussed in Chinese and/or English. However, most results and findings can be applied to other pro-drop languages, which is further supported by other works~\cite{ri2021zero,aloraini2020cross,vincent2022controlling}. In Appendix \S\ref{app:a-1}, we add details on the phenomenon in various pro-drop languages such as Arabic, Swahili, Portuguese, Hindi, and Japanese.

    \item[2.] {\em More Details on Datasets and Methods}: We have no space to give more details on datasets and models. We will use a Github repository to release all mentioned datasets, code, and models, which can improve the reproducibility of this research direction.
\end{itemize}

\section*{Ethics Statement}
We take ethical considerations very seriously, and strictly adhere to the ACL Ethics Policy.
In this paper, we present a survey of the major works on datasets, approaches and evaluation metrics that have been undertaken in ZPT.
Resources and methods used in this paper are publicly available and have been widely adopted by researches of machine translation. We ensure that the findings and conclusions of this paper are reported accurately and objectively.

\bibliography{anthology,custom}
\bibliographystyle{acl_natbib}

\clearpage

\appendix
\section{Appendix}
\label{sec:appendix}

\subsection{Zero Pronoun in Different Languages}
\label{app:a-1}

The pronoun-dropping conditions vary from language to language, and can be quite intricate. Previous works define these typological patterns as pro-drop that can be subcategorized into three categories (as shown in Figure~\ref{fig:zp_overview}): 
\begin{itemize}[leftmargin=*,topsep=0.1em,itemsep=0.1em,parsep=0.1em]
\item {\em Topic Pro-drop Language} allows referential pronouns to be omitted, or be phonologically null. Such dropped pronouns can be inferred from previous discourse, from the context of the conversation, or generally shared knowledge.

\item {\em Partial Pro-drop Language} allows for the deletion of the subject pronoun. Such missing pronoun is not inferred strictly from pragmatics, but partially indicated by the morphology of the verb. 
\item {\em Full Pro-drop Language} has rich subject agreement morphology where subjects are freely dropped under the appropriate discourse conditions.
\end{itemize}

\subsection{Analysis of Zero Pronoun}
\label{app:a-2}

\begin{CJK}{UTF8}{gbsn}
As shown in Table~\ref{tab:zp-domain}, 26\% of Chinese pronouns were dropped in the dialogue domain, while 7\% were dropped in the newswire domain. 
ZPs in formal text genres (e.g. newswire) are not as common as those in informal genres (e.g. dialogue), and the most frequently dropped pronouns in Chinese newswire is the third person singular 它 (``it'')~\cite{Baran2012AnnotatingDP}, which may not be crucial to translation performance. 

\end{CJK}

\begin{table}[h]
\centering
\scalebox{0.9}{
\begin{tabular}{r rrrr}
\toprule
\bf Genres & \bf Sent. & \bf ZH Pro. & \bf EN Pro. & \bf ZPs \\
\midrule
Dialogue & 2.15M & 1.66M & 2.26M & 26.55\% \\
News & 3.29M & 2.27M & 2.45M & 7.35\% \\ 
\bottomrule
\end{tabular}}
	\caption{\label{tab:zp-domain} Extent of pronoun-dropping in different genres. The {\em Dialogue} corpus consists of subtitles in Opensubtitle2018 and the {\em News} corpus is CWMT2013 news data.}
\end{table}

\subsection{The Linguistic Concept}
\label{app:a-3}
Zero anaphora is the use of an expression whose interpretation depends specifically upon antecedent expression. The anaphoric (referring) term is called an anaphor. Sometimes anaphor may rely on the postcedent expression, and this phenomenon is called cataphora. Zero Anaphora (pronoun-dropping) is a more complex case of anaphora. In pro-drop languages such as Chinese and Japanese, pronouns can be omitted to make the sentence compact yet comprehensible when the identity of the pronouns can be inferred from the context. These omissions may not be problems for our humans since we can easily recall the missing pronouns from the context. 

\subsection{Human Evaluation Guideline}
\label{app:a-4}

{We carefully design an evaluation protocol according to error types made by various NMT systems, which can be grouped into five categories: 1) The translation can not preserve the original semantics due to misunderstanding the anaphora of ZPs. Furthermore, the structure of translation is inappropriately or grammatically incorrect due to incorrect ZPs or lack of ZPs; 
2) The sentence structure is correct, but translation can not preserve the original semantics due to misunderstanding the anaphora of ZPs;  
3) The translation can preserve the original semantics, but the structure of translation is inappropriately generated or grammatically incorrect due to the lack of ZPs;  
4) where a source ZP is incorrectly translated or not translated, but the translation can reflect the meaning of the source; 
5) where translation preserves the meaning of the source and all ZPs are translated. Finally, we average the score of each target sentence that contains ZPs to be the final score of our human evaluation.} For human evaluation, we randomly select a hundred groups of samples from each domain, each group contains an oracle source sentence and the hypotheses from six examined MT systems. Following this protocol, we asked expert raters to score all of these samples in 1 to 5 scores to reflect the quality of ZP translations. For the inter-agreement, we simply define that a large than 3 is a good translation and a bad translation is less than 3. The annotators reached an agreement of annotations on 91\% (2750 out of 3000) samples. 
In general, the process of manual labeling took five professional annotators one month in total, which cost US \$5,000.

\end{document}